# Deep Versus Wide Convolutional Neural Networks for Object Recognition on Neuromorphic System

**Md Zahangir Alom, Theodore Josue, Md Nayim Rahman, Will Mitchell, Chris Yakopcic, and Tarek M. Taha**
Department of Electrical and Computer Engineering, University of Dayton, Dayton, OH 45469, USA.
Email: {alomm1, tjosue1, rahmanm12, mitchellw2, cyakopcic1, ttaha1}@udayton.edu

*Abstract*— **In the last decade, special purpose computing systems, such as Neuromorphic computing, have become very popular in the field of computer vision and machine learning for classification tasks. In 2015, IBM's released the TrueNorth Neuromorphic system, kick-starting a new era of Neuromorphic computing. Alternatively, Deep Learning approaches such as Deep Convolutional Neural Networks (DCNN) show almost human-level accuracies for detection and classification tasks. IBM's 2016 release of a deep learning framework for DCNNs, called Energy Efficient Deep Neuromorphic Networks (Eedn). Eedn shows promise for delivering high accuracies across a number of different benchmarks, while consuming very low power, using IBM's TrueNorth chip. However, there are many things that remained undiscovered using the Eedn framework for classification tasks on a Neuromorphic system. In this paper, we have empirically evaluated the performance of different DCNN architectures implemented within the Eedn framework. The goal of this work was discover the most efficient way to implement DCNN models for object classification tasks using the TrueNorth system. We performed our experiments using benchmark data sets such as MNIST, COIL-20, and COIL-100. The experimental results show very promising classification accuracies with very low power consumption on IBM's NS1e Neurosynaptic system. The results show that for datasets with large numbers of classes, wider networks perform better when compared to deep networks comprised of nearly the same core complexity on IBM's TrueNorth system.**

***Keywords*— *Object recognition; Neuromorphic system; Eedn; Deep CNN; TrueNorth.***

## I. INTRODUCTION

We are living in a world consumed by instrumentation that continuously draws data from many kinds of sensors. Nowadays big data is a challenging issue, and we need high performance information processing systems to solve this big data problem. However, a typical high-performance computing (HPC) environment (such as a supercomputing center, or data processing cluster) requires huge amounts of power. Traditional CPUs with multiple processing cores, and larger implementation of Deep Learning (DL) model on Graphic Processing Units (GPU) based computing systems provide state-of-the-art performance, but it consumes a significant amount of power for performing computations. Therefore, different energy efficient and faster computing systems have been developed in the last few years such as field programmable gate arrays (FPGA) [1, 2] and the IBM's Neurosynaptic TrueNorth chip [15-18]. These specialized computing systems have some constrains as well. The constraints are on the number of inputs, memory capacity, and programmability. Mapping big data processing algorithms to these specialized computing systems is one of the most challenging tasks. Among the many available architectures, IBM's TrueNorth (TN) system is one of the first Neuromorphic Chips, which is very efficient in term of power consumption, and with very high throughput [3, 4]. In addition, the MATLAB based corelet programming language was developed, providing a highly scalable objected oriented programming structure [5, 6]. Currently, there are many different applications implemented on IBM's TrueNorth system which have shown promising performance, including object recognition [7], cyber security [8], optimization approaches on the TrueNorth system [9], convolutional sparse coding [10], and many more [11].

Data processing algorithms are always undergoing improvements, and deep learning algorithms have become one of the most prevalent techniques for extracting complex high-level features for object classification and recognition. Deep Learning algorithms, Convolutional Neural Networks (CNN) in particular, use a layered, or hierarchical data representation and learning approach [12, 13]. Furthermore, researchers using modified CNNs have reported improved results for object recognition on different benchmarks including MNIST, CIFAR 10 or 100, Caltech 101 or 256, ImageNet, and many more [ 13, 14]; as well as improved object detection [15] tasks. Accordingly, DCNN approaches have become very popular and widely used in machine learning and computer vision tasks; the main drawback, however, is the increased computational complexity of convolutional network models. In most of the implementations, GPUs are used for training the big networks, which, in most of the cases are utilizing wider and deeper networks for training with higher precision (more than or equal 32 bits) on different benchmarks [16]. As the network size increases, the computational parameters also increase dramatically. It follows that the increased computational costs resulting in significantly greater power consumption due to the use of power hungry GPUs [17].

As the amount of data and data sources are increasing dramatically, deep learning has been playing a key role by providing the solutions for Big Data analytics, data representations, and restoration. In 2015, IBM released the TrueNorth chip, a very low power Neuromorphic processor made up of a massively parallel architecture. TrueNorth is ideally suited to address the big data problem with a significantly lower power profile than conventional systems. Following the trend of deep learning development, IBM released the Energy

Efficient Deep Neuromorphic Network (Eedn) framework for implementing CNN approaches on the TrueNorth system [6]. Accordingly, it becomes very important to implement and evaluate different Deep Learning models for different applications on the very power efficient IBM TrueNorth system. In this implementation, we have implemented different DCNN architectures utilizing the Eedn framework. The contributions of this paper are summarized as follows:

- Implemented different energy efficient DCNN models with Eedn framework.
- Experimented on three popular benchmarks to evaluate different architectures of DCNN including MNIST, COIL-20, and COIL-100.
- Experimented with different deeper and wider deep convolutional networks and discovered the impact of deepness and wideness of networks on the TrueNorth system.

The rest of the paper has been organized in the following way: Section II explains the related works, Section III presents the architecture of the TrueNorth system, Section IV discusses the Eedn framework and implementation details of DCNN on TrueNorth systems. Results and discussions are given in Section V. Conclusions with future directions are made in Section VII.

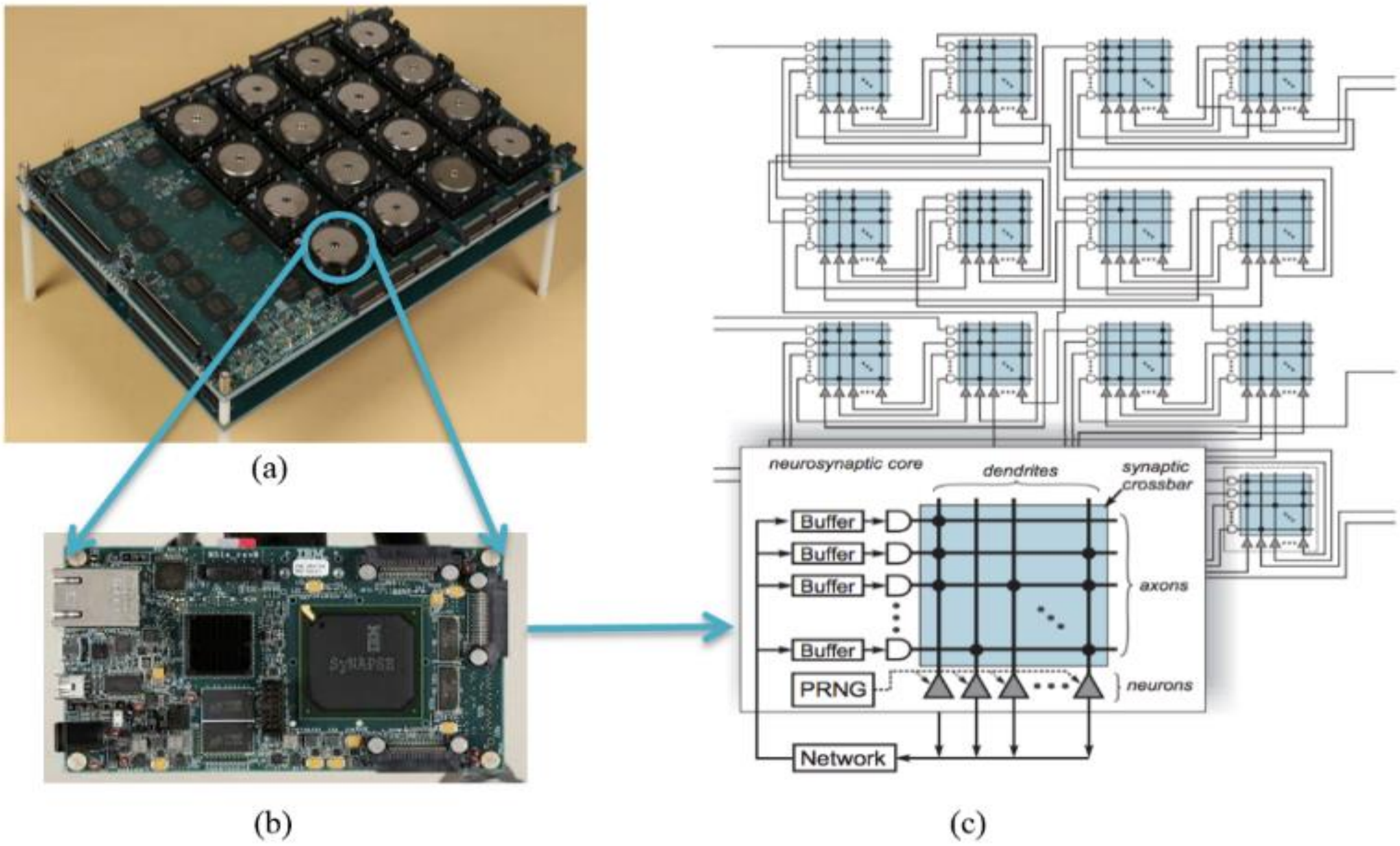

**Fig. 1.** IBM's Neurosynaptic Cognitive TrueNorth Chips: (a) TrueNorth multi-chip system, (b) a single chip, and (c) a zoomed-in internal structure of a core.

## II. Related Works

In the deep learning research community, most of the researcher uses the basic structure of Convolutional Neural Networks (CNNs) with alternative convolution and max-pooling layer followed by a small number of fully connected layers [18]. The piecewise-linear or non-linear activation functions are used within each of these layers. The dropout technique has been used for regularizing the overall network [19]. In addition, drop connection is also used for regularizing the network. However, in general to evaluate those DCNN architectures, the general-purpose computing system such as CPU, GPGPU, and multicore system are used [17]. The optimization of DCNN models are proposed with respect to structural and computational optimization. As structural optimization is of concern, several research studies have been conducted in the community to improve overall accuracy of the DCNN model with lesser numbers of computational parameters, which significantly decreases computational time and power consumption. Some papers have been published on structural optimization of DCNN techniques, which is called SqueezeNet [20]. In most of the cases, these power efficient and faster models are proposed based on low precision implementations of a DCNN [21]. In 2015, Y. Bengio et al., show that a deep network can achieve very high precision training networks using only binary weight values [0 1] [22]. Recently, the ternary weight-based CNN is proposed [23]. The IBM Eedn framework is implemented based on the concept of ternary connected networks [7]. Another controversy is wide versus deep convolutional networks. There are many papers that have been published with full precision implementation on this topic and there is still some research on going.

A recently published paper entitled "Do deep learning need to be deep", clearly stated the impact of network structure on overall recognition accuracy. It concluded that the deeper network (which incorporates more layers for better feature embedding) provides better accuracy compared to the wider network (increase number of neurons with larger number of feature maps in a layer) [24]. In addition, some research was conducted to evaluate the impact of network structure (deeper and wider network) on accuracy with the same number of parameters. This implementation also summarizes that the deeper network performs better compared to wider networks [24]. Another study shows that shallow networks are unable to reach the same levels of accuracy against deep networks with the same number of network parameters. Eventually, they demonstrate that deeper networks provide better performance compared to shallow networks [25].

The question now becomes, is this true in case of the DCNN with the ternary connect method on TrueNorth system? This answer has not yet been determined for IBM's TrueNorth system using a deep learning training methodology with binary weights (0,1), which is well suited to map deep learning onto the TrueNorth system. In 2016, IBM released the Eedn deep learning framework for implementing deep learning on the TrueNorth system, which opened a new opportunity to implement energy efficient deep learning approaches on Neuromorphic hardware [7]. This deep learning framework is very power efficient and provides promising accuracies for image classification tasks. Unlike the implementation of deep learning on CPU, GPGPU, and multicore systems, it is necessary to evaluate the impact of network structures on recognition accuracy for IBM's Neuromorphic TrueNorth system. We have empirically evaluated the performance of different DCNN architectures, tested on different data sets which will help determine efficient design of DCCN models for use on the TrueNorth system; this can lead to the development of additional energy efficient models with better recognition performance.

## III. Neuro-synaptic cognitive system

The traditional von Neumann computing system with a GPU and a multicore processor consumes an abundance of power and area. As systems continue to become larger, the power requirement of these systems have been increasing drastically. To combat this trend, IBM developed and released the TrueNorth Neurosynaptic cognitive architecture as shown in Fig. 2 in 2015. This is an alternative computing system for implementing machine learning, deep learning, and computer vision algorithms with very low power and high energy efficiency [3, 4]. The basic characteristics of IBM's cognitive chip are: first, it is based on a non-von Neumann architecture. Second, it has 4096 cores per chip, each core consists 256 output neurons, each having 256 axons. A 256×256 crossbar of configurable synapses is in each core. Third, each chip contains 1 million programmable neurons and 256 million synapses.

Fig. 1 shows the single chip and multi-chip systems with internal architectural details. The overall TrueNorth architecture is parallel and easily scalable. The internal operation and communication between axons, neurons and other units is performed in spiking form. It is a high throughput Neurosynaptic chip that is capable to run between 1200 and 2600 frames per second using only 25 and 275 $mW$ respectively (effectively greater than 6,000 frames per second per watt) [26]. Each individual axon is assigned 1 of 4 axon types that provides a nine signed bits integer synaptic strength to the corresponding synapse. All event routing inside the chip is completed asynchronously. Each neuron is represented with over 20 individual programmable features, such as synaptic weight, crossbar weight, threshold, leak, and reset. The structure of TrueNorth is very efficient because of the following reasons: first, formation of neurons clusters which is created from inputs of similar pools of axons. Second, spiking events only, which are sparse with respect to time and the communications among the cores performed through a long-distant communication network. Third, the active power of this architecture is proportional to the firing activity.

### A. *Data encoding on TrueNorth chip*

The human brain works in the spiking form and represents non-binary information as binary spikes [3] There are four type of neural coding schemes defined to represent different types of information in the TrueNorth system. The neural coding schemes are binary code (B), rate code (R), population code (P), and time-to-spike (T) code. In general, for data encoding rate coding scheme is used.

### B. *Neurons*

Different types of neuron models can be modeled in the TrueNorth system. For purposes of this study, the Leaky Integrate-and-Fire (LIF) neuron model will be examined. The following five basic operations describe the LIF neuron model: synaptic integration, leak integration, threshold, spike firing, and reset. In general cases, the LIF neuron model can be described by the following equations:

Synaptic integration:

$$V_j(t) = V_j(t-1) + \sum_{i=0}^{N-1} x_i(t)\, s_i \qquad (3)$$

Leak integration:

$$V_j(t) = V_j(t) - \lambda_j \qquad (4)$$

Threshold, fire, and reset

$$\text{If} \quad V_j(t) \geq \alpha_j \qquad (5)$$

$$V_j(t) = 1$$

Else

$$V_j(t) = 0$$

End-if

The parameter $V_j(t)$ in the above equations stands for the sum of membrane potential of the $j^{th}$ neuron in the $t^{th}$ timestep, and $V_j(t-1)$ is the sum membrane potential of the previous timestep. $x_i(t)$, and $s_i$ are the synaptic input as sum of spike input in the current time step and the signed synaptic weights respectively. $\lambda_j$ is the leak value that is subtracted in every time-step from membrane potential. Then the membrane potentials are compared with the threshold voltage $\alpha_j$. If the membrane potential is greater than or equal to the threshold voltage, the neuron fires a spike and resets the membrane potential.

### C. *Crossbar Weights and Synaptic weight*

The crossbar weights $w_{i,j} \in \{0,1\}$ of the neurosynaptic core are 0 or 1 (representing active or inactive states) and are represented using a single bit per weight. Moreover, each active synapse can have one of the four values as its synaptic weight $s_j^{G_i}$ depending on the axon type. There are four types of axons which are determined with the values of {0 1 2 3}. In this work, the default value of $S_j$ {8 4 2 1} are used as synaptic weights [3].

## IV. IMPLEMENTATION WITH ENERY EFFICIENT DEEP NETWORKS (EEDN)

### A. Eedn

IBM's Eedn is a complete deep learning framework for the TrueNorth Neuromorphic system that is used for training, mapping the network onto the Neuromorphic chip, and testing of a DCNN model. This framework consists of different types of layers to construct deep convolutional architectures. The layers are: input layer, pre-processing layer, convolution layer, network in network layer, pooling layer, and drop out layer. During the training, the convolution layer performs basic convolutional operation respect to filter with input features of this layer. For example: if the input dimension is 28x28 and filter size is 3x3 with 12 feature maps then the output size of this layer will be 26x26x12. The pooling or sub-sampling operation is performed using convolution with stride size 2. The dropout operations are applied with fractional value and we have used 0.5 in this implementation. In the training phase, deep neural network is trained using some steps on the GPU which is given in Algorithm I [7]:

**Algorithm I.** Training steps for DCNN on TrueNorth

**Step 1.** Training performs iteratively
**Step 2.** The network's response is computed through the forward pass of network
**Step 3.** The network errors are calculated with network outputs and desired outputs
**Step 4.** The gradient errors are computed at each synapse in the backward pass
**Step 5.** Update weight along with gradient respect to the errors

After successfully completing the training process, the network is mapped onto the IBM's TrueNorth system. In this case, the grouping approach is used in the convolution layers. Let's considered the following components such as mask size or kernel size (K), number of features map (F), and the number of group (G). However, during the implementation of the network onto the TrueNorth system, the following conditions need to be satisfied: first, the number of inputs must be less than or equal to 128.

$$K \times K \times \frac{F}{G} \leq 128 \qquad (6)$$

Second, the number of group of $i^{th}$ layer must divisible with the number of feature maps of $i-1^{th}$ layer.

$$G_i^n \ \% \ F_{i-1}^n = 0 \qquad (7)$$

In Eq. 7, $G_i^n$ is the number of group of $i^{th}$ layer and $F_{i-1}^n$ is number of feature maps of $i-1^{th}$ layer. Third, the total number of cores (TNC) of the architecture must be less or equal to 4096 cores.

$$TNC \leq 4096 \qquad (8)$$

### B. DCNN on TrueNorth:

In this paper, we have empirically evaluated the performances of different architecture of DCNN on different benchmarks. The DCNN architecture consist of different components including: preprocessing layer (P), Convolution layer (C), sub-sampling layer (S), and Network in Network (NiN) layers. We have tested different networks; however, the deeper and wide network structure are shown in Fig. 2 and Fig. 3 respectively. Here one example network architecture is provided for deep and wide network which contains 4064 and 4096 cores respectively. Due to the different hyper parameter of Eedn including number of feature maps and number of groups, it is hard to implement a network model with same number of cores. For example, if case of first implementation of deep model the number of splitter is 384 which is used in the second convolution layer in C2. On the other hand, the total number of splitter is 384+1024 = 1408 which is used in C2 and C4 respectively.

| Lyr | Layer Size Row × Col × Ftr (Grp) | Patch Str | Patch Row Col Ffr | TN Cores Base+Splt | Patch-in-Image Str | Patch-in-Image Row × Col |
|---|---|---|---|---|---|---|
| I | 32 × 32 × 3 | | | | | |
| P1 | 32 × 32 × 12 (1 ) | 1 | [3 × 3 × 3 ] | 0 +0 | 1 | [3 × 3 ] |
| C2 | 16 × 16 × 252 (2 ) | 2 | [4 × 4 × 6 ] | 512 +384 | 2 | [6 × 6 ] |
| C3 | 16 × 16 × 256 (2 ) | 1 | [1 × 1 × 126] | 512 +0 | 2 | [6 × 6 ] |
| C4 | 8 × 8 × 256 (8 ) | 2 | [2 × 2 × 32 ] | 512 +0 | 4 | [8 × 8 ] |
| C5 | 8 × 8 × 512 (32 ) | 1 | [3 × 3 × 8 ] | 512 +0 | 4 | [16 × 16 ] |
| C6 | 8 × 8 × 512 (4 ) | 1 | [1 × 1 × 128] | 256 +0 | 4 | [16 × 16 ] |
| C7 | 8 × 8 × 512 (4 ) | 1 | [1 × 1 × 128] | 256 +0 | 4 | [16 × 16 ] |
| C8 | 8 × 8 × 512 (4 ) | 1 | [1 × 1 × 128] | 256 +0 | 4 | [16 × 16 ] |
| C9 | 4 × 4 × 512 (16 ) | 2 | [2 × 2 × 32 ] | 256 +0 | 8 | [20 × 20 ] |
| C10 | 4 × 4 × 1024 (64 ) | 1 | [3 × 3 × 8 ] | 256 +0 | 8 | [36 × 36 ] |
| C11 | 4 × 4 × 1024 (8 ) | 1 | [1 × 1 × 128] | 128 +0 | 8 | [36 × 36 ] |
| C12 | 2 × 2 × 1024 (32 ) | 2 | [2 × 2 × 32 ] | 128 +0 | 16 | [44 × 44 ] |
| C13 | 2 × 2 × 1024 (8 ) | 1 | [1 × 1 × 128] | 32 +0 | 16 | [44 × 44 ] |
| C14 | 2 × 2 × 1024 (8 ) | 1 | [1 × 1 × 128] | 32 +0 | 16 | [44 × 44 ] |
| C15 | 2 × 2 × 2040 (8 ) | 1 | [1 × 1 × 128] | 32 +0 | 16 | [44 × 44 ] |

**Fig. 2**. Deeper network architecture with 15 layers.

| Lyr | Layer Size Row × Col × Ftr (Grp) | Patch Str | Patch Row Col Ffr | TN Cores Base+Splt | Patch-in-Image Str | Patch-in-Image Row × Col |
|---|---|---|---|---|---|---|
| I | 32 × 32 × 3 | | | | | |
| P1 | 32 × 32 × 12 (1 ) | 1 | [3 × 3 × 3 ] | 0 +0 | 1 | [3 × 3 ] |
| C2 | 16 × 16 × 252 (2 ) | 2 | [4 × 4 × 6 ] | 512 +384 | 2 | [6 × 6 ] |
| C3 | 16 × 16 × 512 (2 ) | 1 | [1 × 1 × 126] | 512 +0 | 2 | [6 × 6 ] |
| C4 | 8 × 8 × 512 (32 ) | 2 | [2 × 2 × 16 ] | 1024+1024 | 4 | [8 × 8 ] |
| C5 | 4 × 4 × 1024 (32 ) | 2 | [2 × 2 × 16 ] | 256 +0 | 8 | [12 × 12 ] |
| C6 | 4 × 4 × 1024 (8 ) | 1 | [1 × 1 × 128] | 128 +0 | 8 | [12 × 12 ] |
| C7 | 2 × 2 × 2048 (32 ) | 2 | [2 × 2 × 32 ] | 128 +0 | 16 | [20 × 20 ] |
| C8 | 2 × 2 × 2048 (32 ) | 1 | [1 × 1 × 64 ] | 64 +0 | 16 | [20 × 20 ] |
| C9 | 2 × 2 × 4096 (64 ) | 1 | [1 × 1 × 32 ] | 64 +0 | 16 | [20 × 20 ] |

**Fig. 3.** Wider network model with 9 layers.

## V. EXPERIMENTAL RESULTS AND DISCUSSION

The entire experiments have been conducted on a desktop computer with an Intel ® Core ™ 2 Duo CPU E86 @ 3.33 GHz processor and 12GB of RAM to evaluate the processing time in MATLAB (R2015a). The datasets details are given in the following database section. The model is implemented in MATLAB, using the integrated programming environment called corelet programming for IBM's Neurosynaptic system. There are two platforms to evaluate, first simulation platform called Neurosynaptic Simulator for Corelet System (NSCS) and another is in actual hardware. We have tested this experiment on both platforms. The running environment (platform) can be selected by changing the mode parameters of "TN" and "NSCS". Exact same program can be run on actual TrueNorth chip or simulator depending upon flag "TN" or "NSCS" respectively. It is noted that the outputs of both environment are almost identical. However, in this implementation, we have evaluated the performance on a single chip TrueNorth system.

### *A. Database:*

Three popular benchmarks for digit and object recognition such as MNIST, COIL-20, and COIL-100 datasets are used in this implementation.

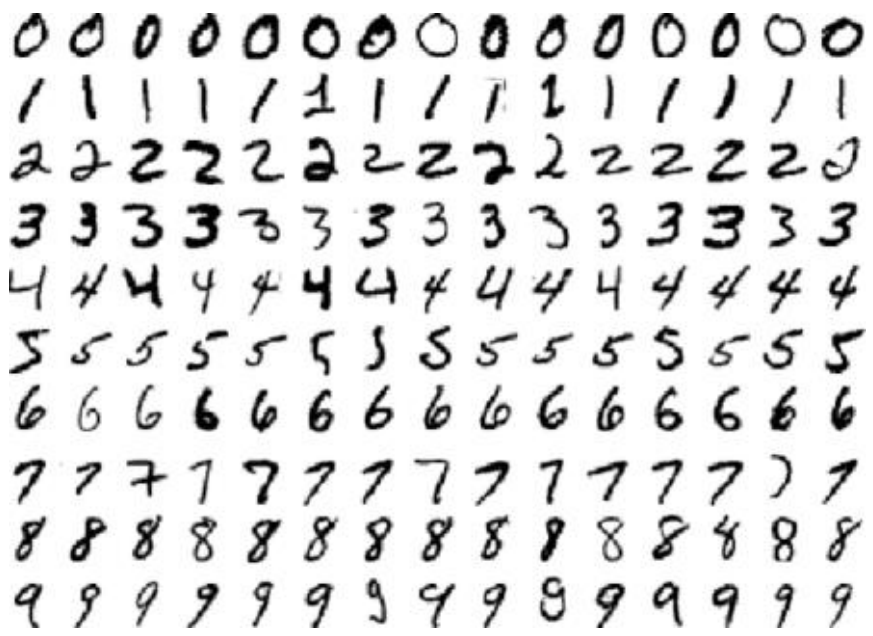

**Fig. 4**. Example samples from MNIST database

#### *1) MNIST:*

MNIST is a one of the benchmark image classification database [27]. This dataset consists with 60000 training samples and 10000 test samples 28x28 gray-scale image representing digits ranging from 0 to 9. We did not apply any data-augmentation except resizing input sample of dataset during this experiment. The samples images are given in below Fig. 4.

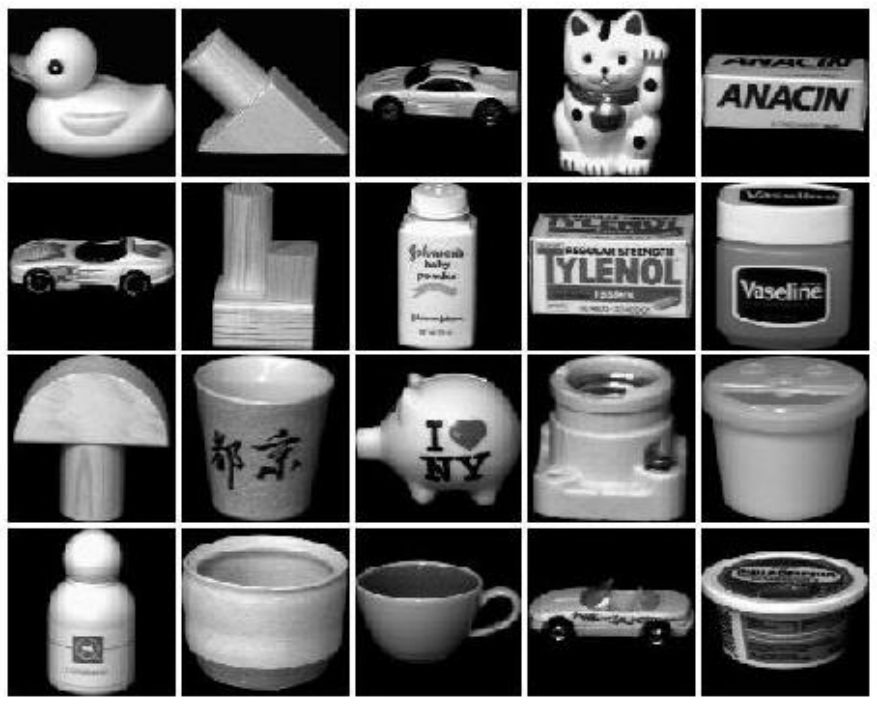

**Fig. 5.** Example images for COIL-20

#### *2) COIL-20 dataset:*

There are two version of database available for Columbia Object Image Library (COIL)-20, the first version of this dataset with background and another version is without background. In this implementation we have used the training and testing samples with background. This database contains 1440 observations (20 objects with 72 poses each) in total, where 1100 samples are used for training the network and remaining 300 samples are used for testing [28]. The example images are shown in the following Fig. 5.

#### *1) COIL-100*

COIL-100 dataset is extended dataset of COIL-20. This dataset contains color images for 100 classes of object. This dataset contains 7000 Color images where 5000 samples are used for training and remaining 2000 samples are used for testing in this implementation. The turntable was rotated through 360 degrees to vary object pose with respect to a fixed color camera. Images of the objects were taken at pose intervals of 5 degrees. This corresponds to 72 poses per object [29]. Due to the input size contained of TrueNorth system, we have resized the input sample to 32x32 pixels. The examples images of dataset are shown in Fig. 6.

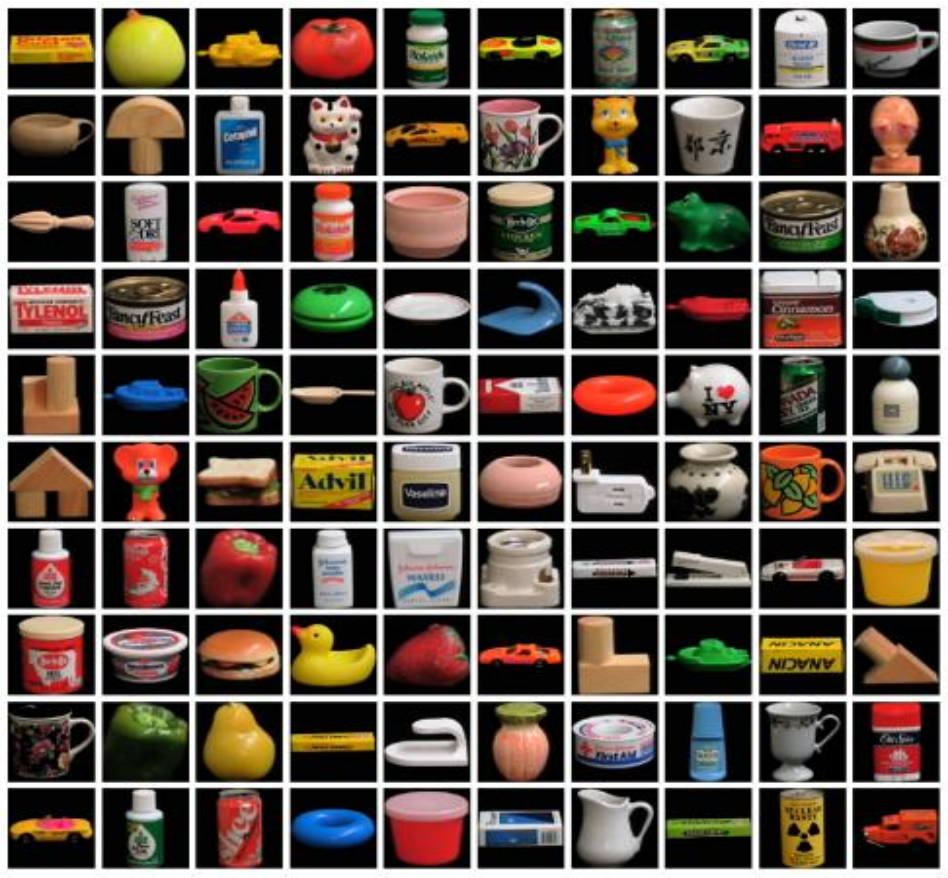

**Fig. 6**. Example image of COIL-100 dataset.

### *B. Results*

We have evaluated the performance of different architectures which consist of different numbers of layers and cores on TrueNorth system. The performance of network varies with respect to the number of cores. We have tested on MNIST, COIL-20, and COIL-100 datasets. We have also investigated the variation of recognition accuracy respect structure of network with same number of cores on IBM's TrueNorth.

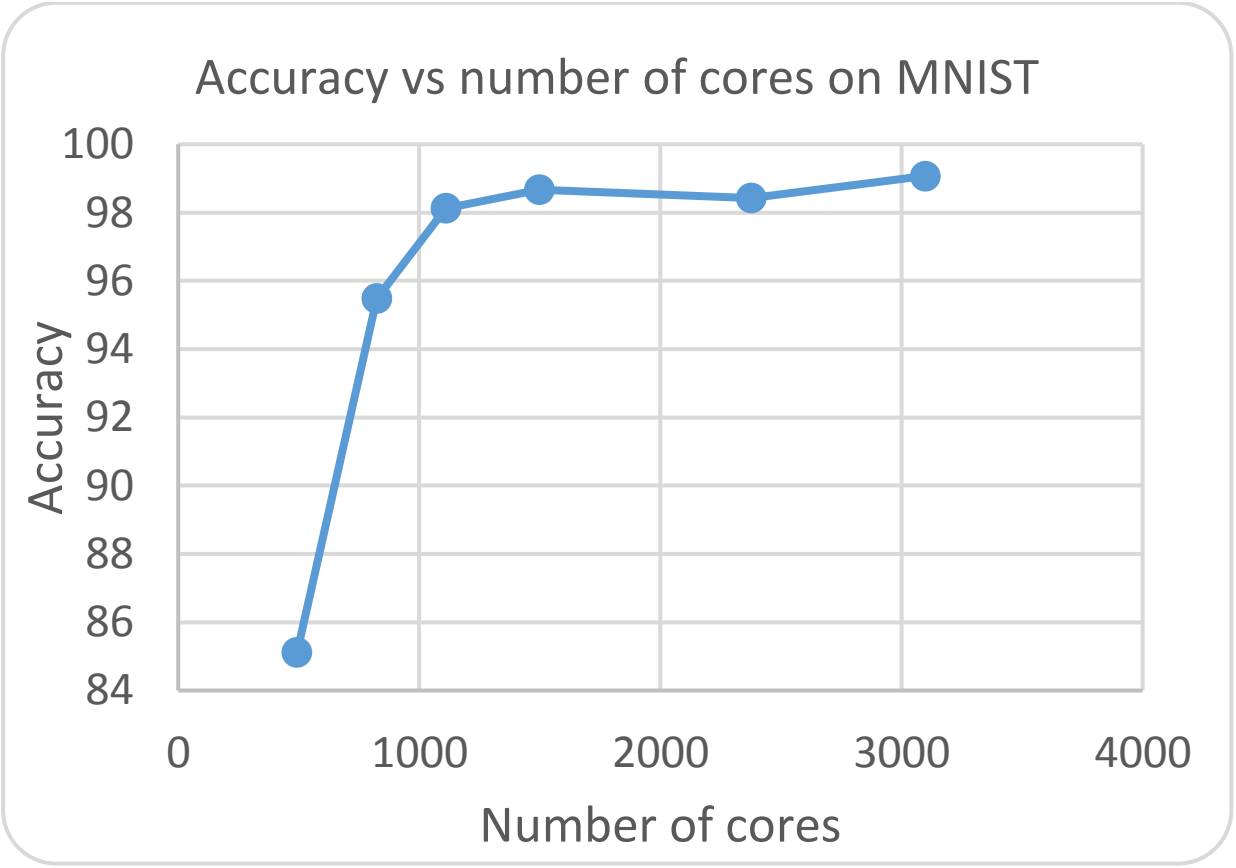

**Fig. 7.** Testing accuracy versus number of cores on MNIST dataset.

*1) MNIST:*

To evaluate the performance on MNIST dataset, we have taken the default implementation network for MNIST dataset in Eedn framework and we have varied number of features maps and groups and testing with different architectures. Fig. 9 shows the accuracy with respect to the number of cores with different architecture. The figure clearly shows that the performance increase with respect to the number of cores used with bigger networks. Fig. 7 shows that testing accuracy for MNIST dataset with a network consisted with more cores with bigger structure and we have achieved around 99.07 percent accuracy with deeper network. In addition, we have also tested with wide version the same network with almost same number of cores. However, from Fig. 8, it can be clearly concluded that the deeper network provides better testing accuracy compare against wider network.

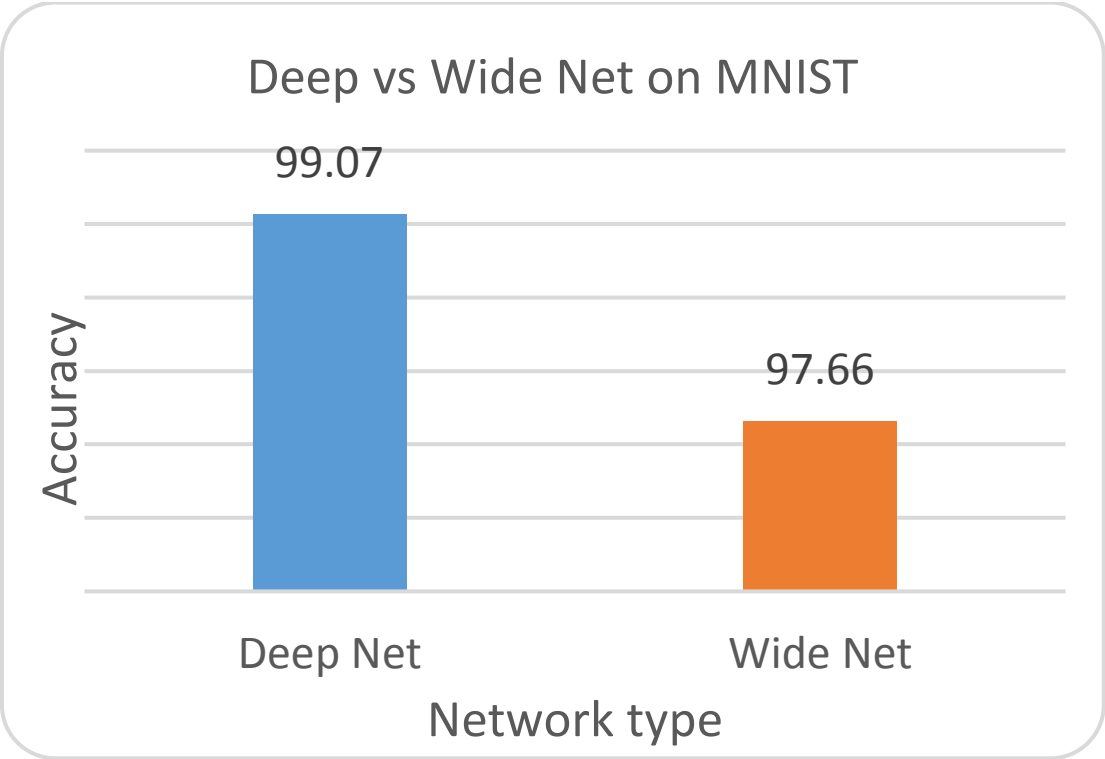

**Fig. 8**. Comparison of testing accuracy of deep versus wide network on MNIST dataset.

*2) COIL-20*

The following figures show the training loss and accuracy of this implementation for 3000 iterations. From the Fig. 9, it is clearly shown that DCNN model on TrueNorth system provides promising recognition accuracy with only 3000 iterations on COIL-20 dataset. After round 1000 iteration, we have achieved almost 100% training accuracy on this dataset.

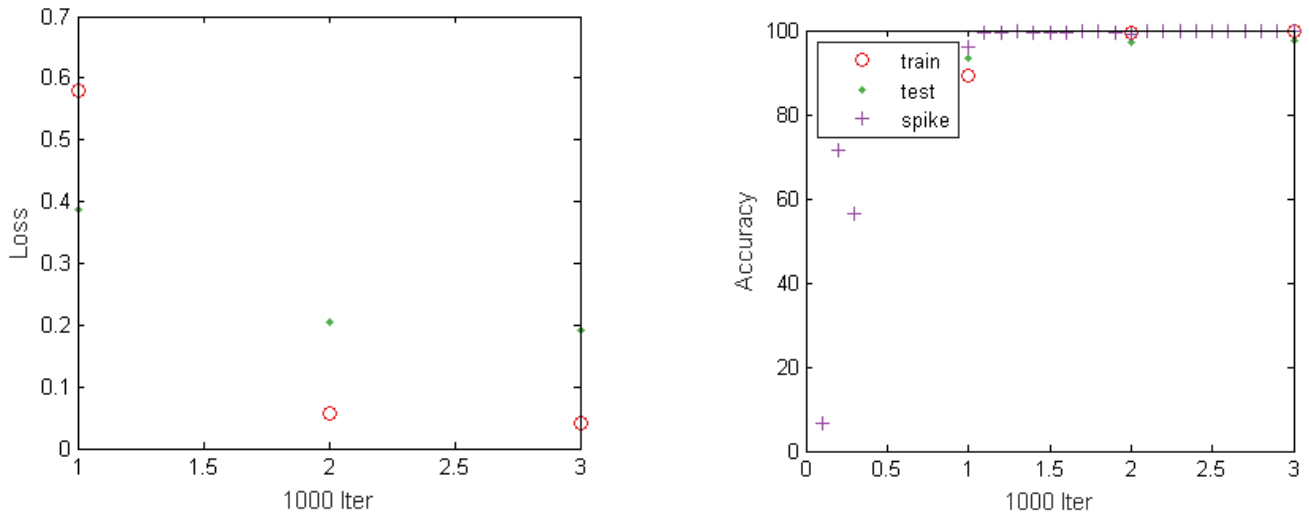

**Fig. 9**. Training loss for COIL-20 only for 3000 iterations

We have also investigated different models on TrueNorth system with different models with different number of cores from 480 to 4094 cores shown in Fig. 10. According to the Fig. 10, we have observed the highest accuracy with only around 1400 cores. The very close accuracy is observed with 4064 cores on TrueNorth for COIL-20 dataset. It is noted that for all different networks, we have conducted experiment with only 3000 iterations. In both experiment of COIL-20 and COIL-100, we have used batch size 50 and learning rate 0.1 and 0.01.

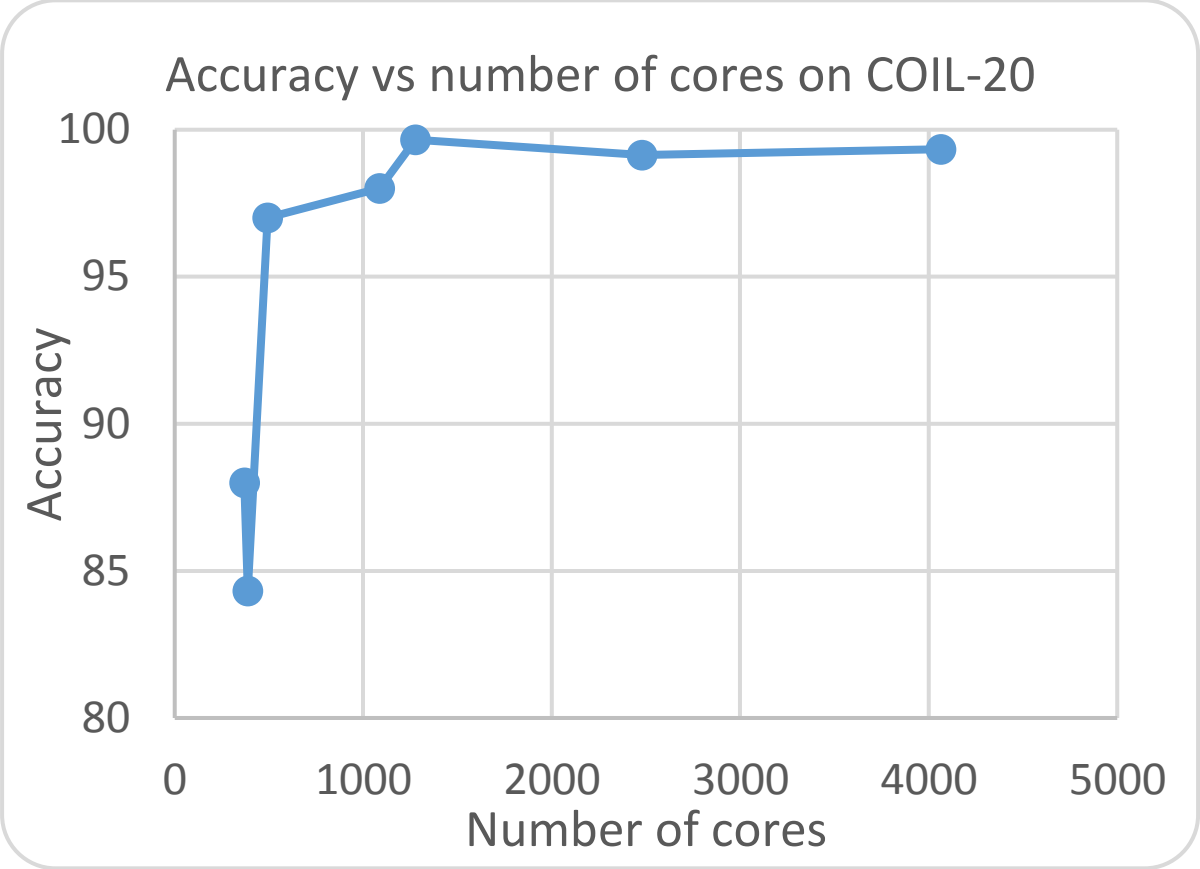

**Fig. 10.** Network accuracy respect to the number of cores on COIL-20 dataset.

However, we have implemented two version of the network with 4064 and 4096 cores respectively; one is deeper (increase number of layers within network) and another one is wider (increase number of neurons in the network). Fig. 11 shows the recognition accuracy for COIL-20, where deep networks uses 4064 cores and wide version of network utilizes 4096 cores on single chip implementation which are shown in Fig. 2 and Fig. 3 respectively. Fig.11 shows the testing recognition accuracy for COIL-20, the wider version networks provide around 99.36% recognition accuracy whereas the deeper of the network shows 99.23% accuracy. According to the Fig. 11, it is clearly concluded that the wider network provides the better recognition accuracy compare to the deeper network with almost same number of cores.

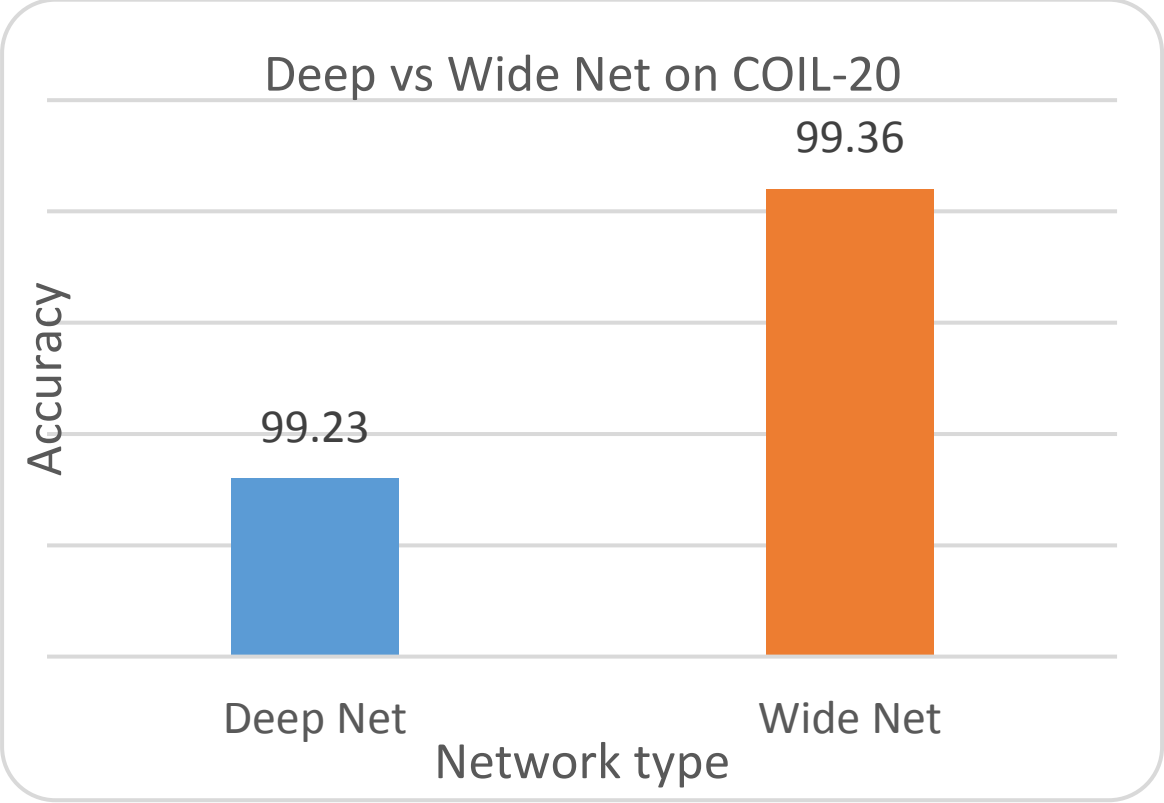

**Fig. 11.** Testing with respect to deep versus wide network on COIL-20

*3) COIL-100*

The training loss and training accuracy are shown in Fig. 12. Fig. 12(a) shows the loss for training. According to the figure, it can be said that the convergence of the network during training is fast. Fig. 12(a) shows the training loss for 30,000 iteration and Fig. 12(b) shows the training and testing accuracy with red and green color respectively.

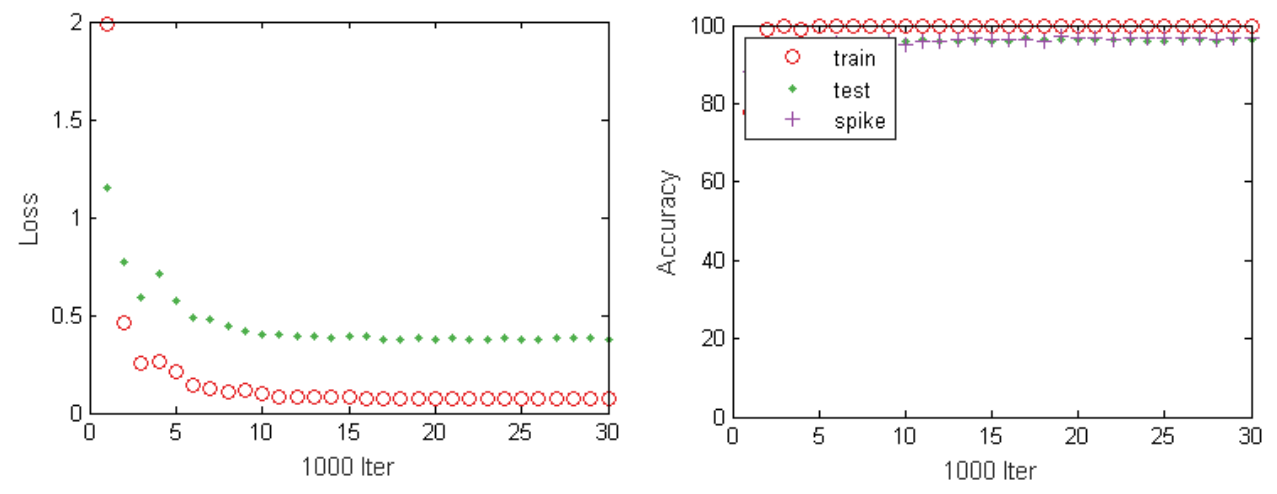

(a) (b)

**Fig. 12.** Training loss and accuracy for COIL-100 dataset: (a) Training loss and (b) Training and testing accuracy.

The weights updating status during training are shown in Fig. 13.

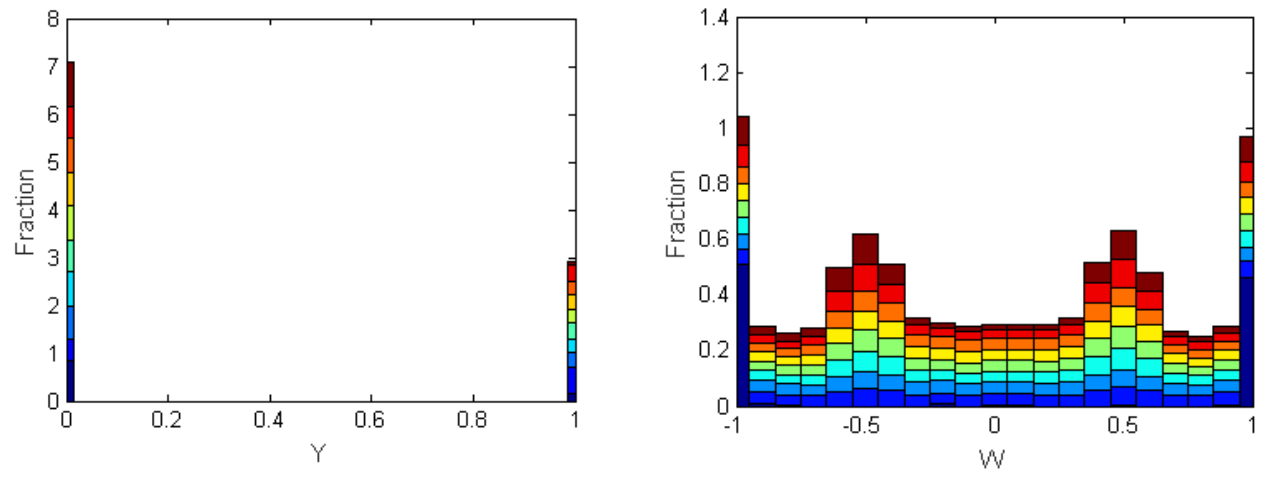

**Fig. 13.** Weight update status during training.

We have also conducted the experiment with different wide and deep networks on COIL-100 for 30000 iterations. As with COIL-20, we have investigated different network with 520, 840, 2200, 4064, and 4094 cores. The experiment results are shown in Fig. 14, and it is clearly shown that the bigger network with more cores perform better compared to the smaller network. We have achieved the best accuracy with the biggest network 4096 cores.

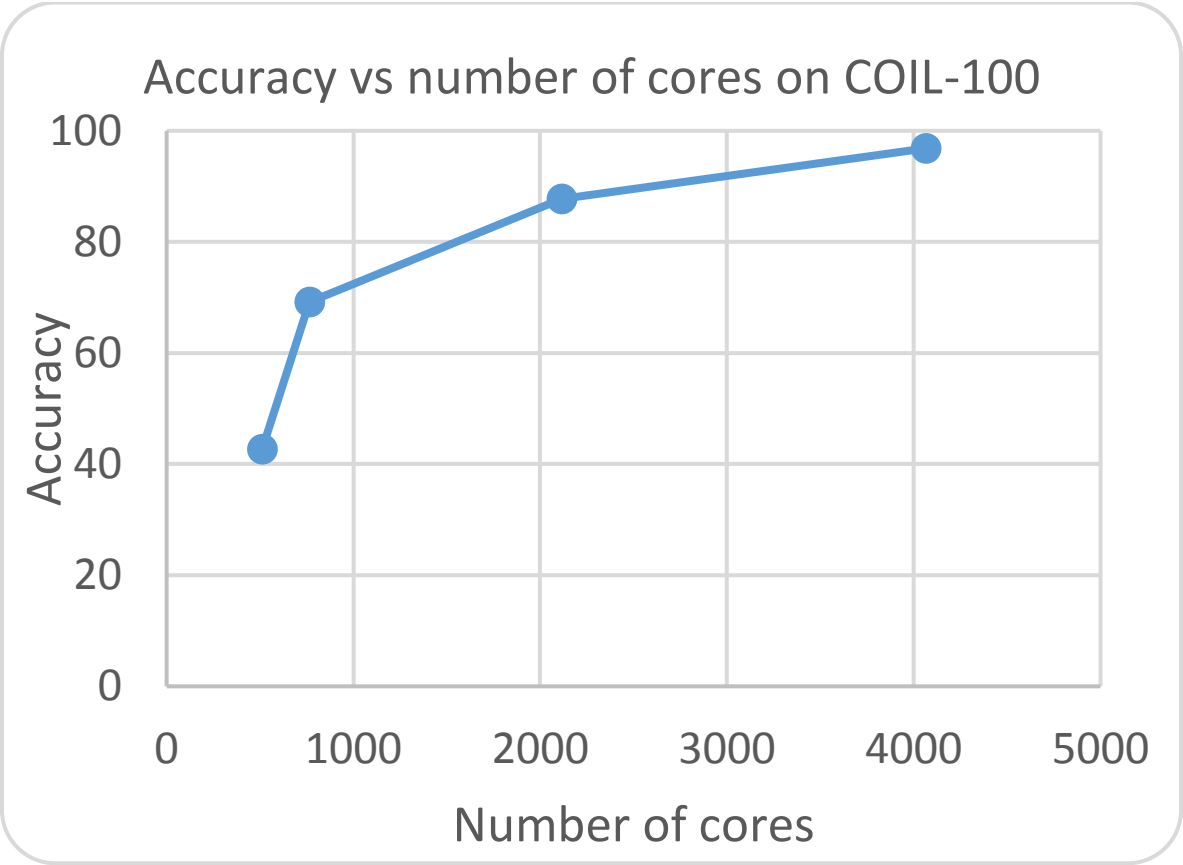

**Fig. 14.** Testing accuracy versus number of cores on COIL-100 dataset.

The experimental results for deeper versus wider networks is shown in Fig. 15. The deeper network contained 15 layers and 4064 cores and wider version of network contained 8 layers with 4096 cores. The architectures of the deeper and wider networks are shown in Fig. 2 and Fig. 3 respectively. The wider network shows better results compare to deeper network in this case. The result shows around 96.8% testing accuracy on both simulator and TrueNorth chip.

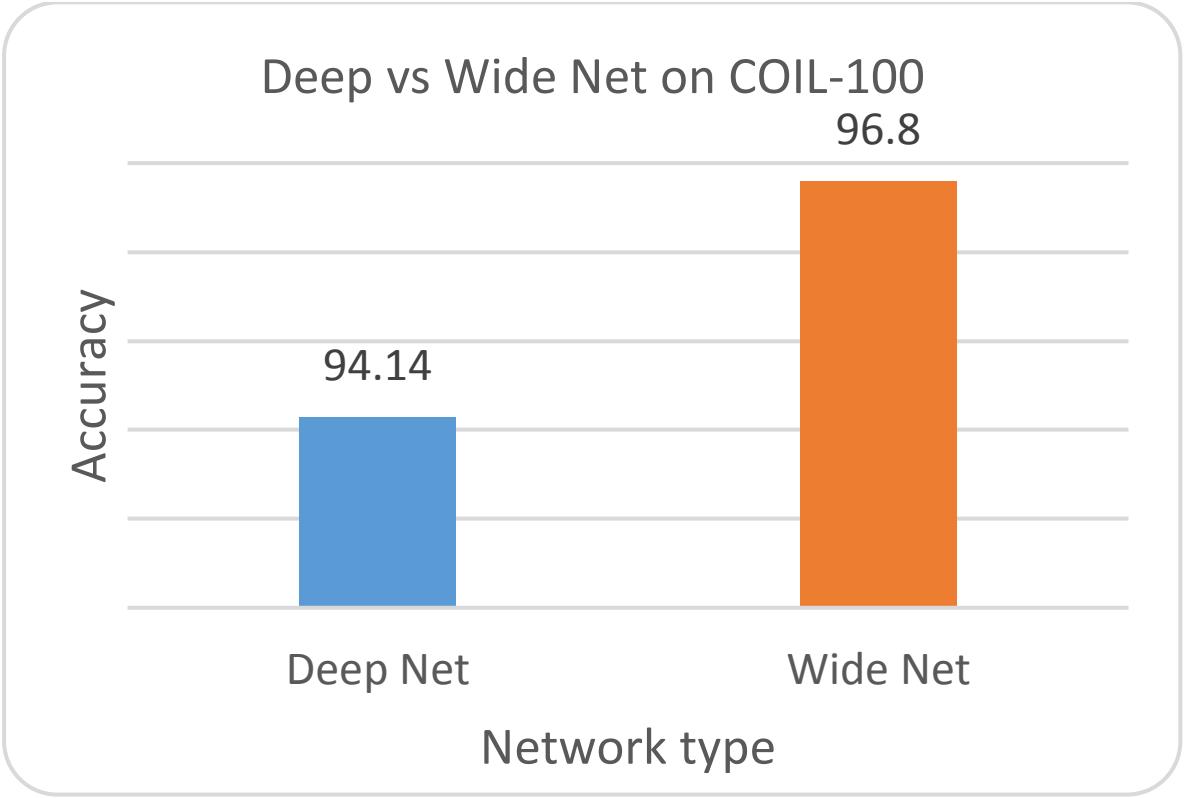

**Fig. 15.** Deeper versus wider network on COIL-100

## C. *Evaluation*

When desiring to utilize more cores in IBM's TrueNorth Neuromorphic system, it is difficult to implement a network that utilizes the maximum number of core resources. We were limited to using a single chip TrueNorth system that contains 4096 cores. However, the architectures we have explored and evaluated in this experiment are limited to 4096 cores. The DCNN architecture on TrueNorth, the layers at the beginning of the network requires more cores whereas the posterior layer needs less cores for mapping onto the chip. For mapping the network onto TrueNorth system, the splitter cores are used. It is observed from the network architecture that the wider network requires more splitter cores (1408) compare to the number of splitter cores (384) of deeper network. From the experimental results, it is clearly observed that the recognition accuracy varies with respect to the network architecture and number of cores on TrueNorth system. As the number of classes increase, the wider network performs better compare against deeper network with almost same number of cores which are evaluated with a set of experiment.

In addition, we have implement DCNNS model with Keras and TensorFlow on the back end on a single GPU machine. This model consisted with seven layers including Softmax layer and contains around 0.5 million network parameters and training with ADAM optimizer with learning rate 0.001. We have achieved 100% accuracy for both COIL-20 and COIL-100 dataset. On the other hand, we have achieved 99.36% testing accuracy for COIL-20 and around 97% testing accuracy for COIL-100 on IBM's TrueNoth system. Although we have received 0.64% and around 3% less testing accuracy, the DCNN models on TrueNorth has significant advantage in term of power.

## D. *Power consumpiton*

The power consumption of TrueNorth system is compared with traditional computing system in this section. Traditional computing systems, such as CPUs and GPUs easily consume 100W or more power, whereas an entire TrueNorth system consumes only up to 100mW to operate the 4096 cores on it. Here, we have used almost all cores for object recognition task with different network architectures. In addition, it is noted that about 50% of the power is passive power in TrueNorth system. The overall power requirement for 4096 cores is $100\, mW$. However, for 4064 cores of the deeper network, $0.5 * 100mW + (4064/4096) * 50\, mW = 99.64\, mW$ is required. We have achieved almost the same level of recognition accuracy which are achieved by CPU and GPU system for all datasets. However, implementation on the TrueNorth system requires significantly lower power with respect to traditional computing system.

## VI. Conclusion

This work represents a very important step towards evaluation of impact of the architecture of DCNN and number of cores on recognition accuracy in TrueNorth neuromorphic computing system. We have empirically evaluated the recognition accuracy of different DCNN models on three popular benchmarks including MNIST, COIL-20, and COIL 100 on IBM's TrueNorth system. We have achieved about 99.07%, 99.36% and 96.8% as the highest testing accuracy on MNIST, COIL-20 and COIL-100 respectively. The experimental result shows the wider version of the network outperforms the deeper version of the network with the same number of cores on the TrueNorth system. We have achieved the highest accuracy with the wider network for COIL-20 and COIL-100 datasets with almost the same number of cores compared to the deeper network. In the future, we would like to conduct this experiment with more complex datasets on a multi-chip TrueNorth system.

## Acknowledgment

This work is supported by the US National Science Foundation (NSF).